\pdfoutput=1
\documentclass[11pt]{article}

\usepackage{naacl2021}

\usepackage{times}
\usepackage{latexsym}

\usepackage[T1]{fontenc}
\usepackage[utf8]{inputenc}

\usepackage{microtype}

\usepackage{amsthm}
\usepackage{amsfonts}
\usepackage{amsmath}
\usepackage{amssymb}
\usepackage{tikz}
\usepackage{tikz-cd}
\usepackage[ruled,vlined]{algorithm2e}
\usepackage{algpseudocode}

\title{Fabula Entropy Indexing: Objective Measures of Story Coherence}

\newtheorem{theorem}{Theorem}[section]

\makeatletter
\newtheoremstyle{definition}
  {3pt}% space before
  {3pt}% space after
  {\addtolength{\@totalleftmargin}{1em}
   \addtolength{\linewidth}{-1em}
   \parshape 1 1em \linewidth}% body font
  {}% indent
  {\bfseries}% header font
  {.}% punctuation
  {.5em}% after theorem header
  {}% header specification (empty for default)
\makeatother

\theoremstyle{definition}
\newtheorem{definition}{Definition}[section]

\theoremstyle{definition}

\author{Louis Castricato, 
  Spencer Frazier,
  Jonathan Balloch, and 
  Mark O. Riedl\\
  Georgia Tech \\
  \texttt{\{lcastric, sfrazier7, balloch\}@gatech.edu, \{riedl\}@cc.gatech.edu} \\}

\begin{document}
\maketitle
\begin{abstract}
Automated story generation remains a difficult area of research because it lacks strong objective measures. Generated stories may be linguistically sound, but in many cases suffer poor narrative coherence required for a compelling, logically-sound story. 
To address this, we present  Fabula Entropy Indexing (FEI), an evaluation method to  assess story coherence by measuring the degree to which human participants agree with each other when answering true/false questions about stories. 
We devise two theoretically grounded measures of reader question-answering entropy, the entropy of world coherence (EWC), and the entropy of transitional coherence (ETC), focusing on global and local coherence, respectively.
We evaluate these metrics by testing them on human-written stories and comparing against the same stories that have been corrupted to introduce incoherencies.
We show that in these controlled studies, our entropy indices provide a reliable objective measure of story coherence.

% We find that when artificially corrupting our story worlds and plots, these changes can be directly observed quantitatively in a reproducible method.

% Similarly, we devise two separate measures for Fabula Entropy Indexing: Entropy of World Coherency (EWC) and Entropy of Transition Coherency (ETC) which measure story world coherency and plot transitional coherency respectively. We show that both of these methods are robust. 

\end{abstract}

\section{Introduction}

Automated story generation is one of the grand challenges of generative artificial intelligence. 
AI storytelling is a crucial component of the human experience.
Humans have always used storytelling to entertain, share experiences, educate, and to facilitate social bonding. 
For an intelligent system to be unable to generate a coherent story limits its ability to interact with humans in naturalistic ways.

There have been a number of techniques explored for story generation; these include symbolic planning, case-based reasoning, neural language models and others. Despite extensive research, automated story generation remains a difficult task.
% Early story and plot generation systems relied on symbolic planning~\cite{meehan1976metanovel,  lebowitz1987planning,cavazza2003interacting,porteous2009controlling, riedl2010narrative,ware2011cpocl}
% or case-based reasoning~\cite{perez2001mexica,peinado2005creativity,turner2014creative}.
% %Symbolic story planners in particular, use hand-crafted domain knowledge representations.
% An increasingly common machine learning approach to story generation is to use neural language models~\cite{roemmele2016writing,khalifa2017deeptingle,clark2018neural,martin2018event}.
% These techniques have improved with the adoption of Transformer-based 
%  models, such as GPT-2~\cite{radford2019language}. 
% While GPT-2 and similar neural language models are considered highly fluent from a grammatical standpoint.
%---they make statistical choices on how to continue a story based on sampling from a learned distribution $P_\theta(tok_n | tok_{n-k}, ..., tok_{n-2}, tok_{n-1})$.
% Human readers, however, do not perceive the coherence of a narrative as a function of the likelihood of seeing particular words based on the occurrence of previous words.

One of the reasons why automated story generation is such a difficult area of research is due to weak objective validation measures.
Traditional automated measures of natural language quality---perplexity and n-gram based methods such as BLEU~\cite{bleu}---are insufficient in creative generation domains such as story generation.
These metrics assume that generated language can only be good if is resembles testing data or a given target story. 
This precludes the possibility that stories may be good yet be completely novel. 
Indeed, the goal of story generation is usually the construction of novel stories.

In the absence of automated evaluation metrics, the alternative is to use human participant studies.
Human participants, typically recruited via crowdsourcing platforms (e.g Mechanical Turk or Prolific), are asked to read the stories generated by various systems and provide subjective rating or rankings.
Questionnaires may ask participants to rate or rank the overall quality of stories, but may also ask specific questions about features of stories such as fluency or coherence. 
Coherence is particularly difficult feature of stories to measure because the term ``coherence'' can mean different things to different participants.
%\MarkRight{not needed here, discuss in related work}
%This has led some researchers to ask questions about features related to coherence such as ``does the story adhere to a single plot?''~\cite{tambwekar,ammanabrolu:aaai2021}.

In this paper, we introduce a technique for {\bf objective} human participant evaluation, called {\em Fabula Entropy Indexing} (FEI).
FEI provides a structure for metrics that more objectively measure story coherence based on human question-answering.
A {\em fabula} is a narratological term referring to the reader's inferred story world that a story takes place in, whether it be similar to the real world or a fantasy or science fiction world. 
The reader may of course be surprised by certain events but other events may seem implausible or contradictory, thus disrupting coherence.
As they read, humans form cognitive structures to make sense of a story, which in turn can be used to answer simple true/false questions about the story.
As such, an {\em incoherent} story results in readers making random guesses about the answers to these questions.
FEI metrics thus measure the entropy of the answers---how much the answers disagree with each other---which directly correlates with the coherence of the story.
%It doesn't matter what the questions are or what the answers are---we do not require a ground truth---but can merely observe the agreement between human participants when answering questions.

We introduce two such FEI metrics: {\em Entropy of Transitional Coherence} (ETC) and {\em Entropy of World Coherence} (EWC), measuring (respectively) sequential coherence between events in a story, and the internal coherence of the \textit{story world}: the facts about characters, objects, and locations that distinguish a story.
The correlation between human question-answering and these metrics are grounded in narratological\footnote{{\em Narratology} is the study of stories and storytelling.} theories. 

% We provide a set of experiments that validate FEI as an evaluation metric.
To validate the measure, we test our metrics on human-written stories as well as corrupted versions of those stories. For the corrupted stories, we artificially reduce the coherence by altering elements of the story.
We show that FEI metrics evaluate non-corrupted human-written stories as having low entropy and corrupted stories as having higher entropy.

%%%%%%%%%%%%%%%%%%%%%%%%%%%%%%%%%%%%%%%%%

\section{Background and Related Work}

\subsection{Automated Story Generation}

Early story and plot generation systems relied on symbolic planning~\cite{meehan1976metanovel,  lebowitz1987planning,cavazza2003interacting,porteous2009controlling, riedl2010narrative,ware2011cpocl}
or case-based reasoning~\cite{perez2001mexica,peinado2005creativity,turner2014creative}.
%Symbolic story planners in particular, use hand-crafted domain knowledge representations.
An increasingly common machine learning approach to story generation is to use neural language models~\cite{roemmele2016writing,khalifa2017deeptingle,clark2018neural,martin2018event}.
These techniques have improved with the adoption of Transformer-based 
 models, such as GPT-2~\cite{radford2019language}. 
While GPT-2 and similar neural language models are considered highly fluent from a grammatical standpoint.

In these systems, a neural language model learns to approximate the distribution $P_\theta(tok_n | tok_{<n})$ where $\theta$ is the parameters that approximate the pattern of an underlying dataset.
Stories are produced by providing an initial context sequence, then iteratively generating additional tokens by sampling from the distribution. 
When the language model is trained on a corpus of stories, subsets of the generated text tend to also be a story.
%Sometimes generation is done hierarchically~\cite{yao2019plan,ammanabrolu2020story}.
%However, coherence is not guaranteed; statistical sampling from a distribution is not constrained to making logical transitions. Rich relationships that readers make to perceive coherence are not modeled.

One of the reasons why story generation is challenging is because of the strong requirement that stories be {\em coherent}.
Coherence can refer to readability/fluency. 
However, stories also require {\em plot coherence}, which is how well the elements of a plot cohere with each other. 
%For other stories, transitional coherence is necessary -- one event must necessarily occur before another given a story world's constraints.
Studies of human reading comprehension~\cite{trabasso1985causal,graesser91,graesser94} show that humans comprehend stories by tracking the relations between events. 
Reader comprehension studies suggest that readers rely on the tracking of at least four types of relations between events: (1)~causal consequence, (2)~goal hierarchies, (3)~goal initiation, and (4)~character intentions. 
The perceived coherence of a story is a function of the reader being able to comprehend how events correlate to each other causally or how they follow characters' pursuits of implicit goals.

To control the generation and achieve greater coherence, a high-level plot outline can either be generated or given as an input to a language model.~\cite{fan2018hierarchical,peng2018towards,rashkin2020plotmachines,brahman2020modeling}.
These techniques can produce more coherent stories when their guidance forces different parts of the story to appear related or to follow a pattern acceptable to humans.

\citeauthor{tambwekar2018controllable}~\shortcite{tambwekar2018controllable} attempt to train a neural language model to perform goal-based generation.
They fine-tune a neural language model with a policy-gradient reinforcement learning technique that rewards the language model for generating events progressively closer to the goal event.
%This has the benefit of improving readers' perceptions of coherence, but---being based on a language model---does not ensure that any transition from one event to the next will be always be perceived as related.

% The C2PO system~\cite{ammanabrolu2020automated} treats story generation as a bi-directional search from an initial event and a goal event.
% The search space is inferred from the COMET~\cite{comet} commonsense inference model . \Spencer{Explain? COMET? How does it constrain the search space}

%%%%%%%%%%%%%

\subsection{Story Generator Evaluation}

Traditional automated measures of natural language quality such as perplexity or n-gram comparisons (e.g., BLEU) 
are generally considered insufficient for evaluating story generation systems.
Perplexity is the measure of how well a model captures the patterns in an underlying dataset.
Implicit in the notion of perplexity is the belief that the quality of a model is tied to its ability to reconstruct its own data.
However, in automated story generation, stories that are very dissimilar to training and testing data can also be ``good''. 
Likewise, BLEU (and related techniques such as ROGUE and sentence mover techniques~\cite{clark2019sentence}) measure a language model's ability to produce $n$-grams in a specific target sentence, whereas a good story may not resemble a given target story and yet still be coherent.

The gold standard for evaluation of automated story generation systems is to use human participant studies.
Many systems are evaluated with subjective questionnaires in which human participants either rate 
% \Mark{example papers that do this} 
generated stories on a scale, or rank pairs of stories. 
% \Mark{examples of papers that do this}.
Often a single question is asked about overall quality.
Other subjective questions focusing on different story attributes, such as coherence, may be asked as well.
Asking questions about coherence is tricky as participants may have different notions of what coherence might mean, from grammatical notions of coherence to logical story structure.

% \citet{Purdy2018PredictingGS} developed a set of algorithms that would predict how humans will answer subjective questions about coherence, consistency, grammaticality, and narrative productivity.
% The goal of the work was to replace expensive human-participant studies.
% These algorithms were validated against human participant responses to the same stories and found to be reliable predictors.
% However, the measure of coherence was found to be the weakest predictor.

\citet{Purdy2018PredictingGS} introduced a set of subjective questions for human participant studies about global coherence, local consistency, grammaticality, and overall story quality. Algorithms to predict how humans would answer these questions were also introduced. The goal of this work was to reduce reliance on expensive human-participant studies.
One innovation is that they don't directly ask about coherence, which can be an ambiguous term, but instead ask questions such as ``the story appears to be a single plot''.
This set of questions has been used by \citet{tambwekar2019controllable} and \citet{ammanabrolu2020automated}. The algorithms introduced by \citet{Purdy2018PredictingGS} were validated and proven to be reliable predictors but the measure of coherence was shown to be the weakest predictor.

% Prior story evaluation methods, like those seen in \cite{Purdy2018PredictingGS} are subjective in nature and require a reader discern what coherence means when comparing two stories. This work was eventually extended in \cite{ammanabrolu2020automated} and introduced pairwise comparisons of stories -- this eliminates a significant amount of ambiguity when defining story coherence. Particularly, if a reader observes two stories and one makes less sense than the other then they would be more likely to choose the more logical story as coherent.

% Over a large enough set of stories - produced by 2 or more models - this should converge to a simple categorical distribution. Stories from model A are ranked X\% more coherent than stories from model B.

The USER technique, introduced as part of Storium~\citep{akoury2020storium}, is a means of evaluating stories by giving human participants the means to edit a generated story.
They measure the largest subsequence not edited by the author during a story continuation. 
They conclude that their measure is strongly correlated with human evaluation of coherency.

% By comparison, machine-in-the-loop methods like \cite{akoury2020storium} draw inspiration from classical NLG evaluation, like ROUGE-L. These methods measure the largest subsequence not edited by the author during a story continuation. They conclude that their USER (???) measure is strongly correlated with human evaluation of coherency. However, in doing so their control metric is poorly grounded. Humans are often poor judges of coherency unless they are actively forced to process the story in working memory. (Cite a paper rogelio sent us a while ago, cant find it right now).

\citet{li2013story} evaluated their story generation system using an objective human participant study.
They generated stories and then had humans add sentences, delete sentences, or swap sentence orderings.
The number of edits is used to score the story generation system (lower is better).

\citet{riedl2010narrative} also evaluated their story generation system with an objective human participant study based on cognitive science.
They conducted a question-answering protocol to elicit the cognitive model that humans had about the causal relations and goals of characters.
Specifically they constructed a number of questions that the story generation system believed human readers should be able to answer. 
The measure of story quality was the degree to which humans answered the questions the way the algorithm predicted they would.
This technique is the most similar in nature to our proposed measure of coherence;
our technique is mathematically grounded and not tied to any particular way of generating stories.

%%%%%%%%%%%%%%%%%%%%%%%%%%%%%%%%%%%%%%%%%%%%%%%%%%%

\section{Preliminaries}

In this section we review narratological definitions that will be relevant to understanding how to measure the Fabula Entropy Indices.

\begin{definition}
A {\bf narrative} is the recounting of a sequence of events that have a continuant subject and constitute a whole \cite{prince2003dictionary}. 
\end{definition}

\noindent
An event describes some change in the state of the world. 
A ``continuant subject'' means there is some relationship between the events---it is about something and not a random list of unrelated events. 
All stories are narratives, but also include some additional criteria that are universally agreed upon.

% \begin{definition}
% A {\bf story world} is the set of all characters, objects, and locations that can be inferred to exist from reading a story. The story world is assumed distinct from the real world, though it can closely resemble the real world~\Mark{cite: Laure-Ryan}
% \end{definition}

% We will not give a definition of story, as it is the subject of debate with no clear dominant definition. 
% All stories are narratives.
% It is common to propose that a story is distinguished from a narrative in that a story has some specific criteria 
% , but not all narratives are stories. Unfortunately I cannot point to a specific set of criteria that makes people regard a narrative as a story. One strong contender, however, is a structuring of events in order to have a particular effect on an audience.

Structural narratologists suggest there are different layers at which narratives can be analyzed: {\em fabula} and {\em syuzhet} \cite{bal2009narratology}

\begin{definition}
The {\bf fabula} of a narrative is an enumeration of all the events that take place the story world. 
\end{definition}

\begin{definition}
The {\bf syuzhet} of a narrative is a subset of the fabula that is presented via narration to the audience. 
\end{definition}

\noindent
The events in the fabula are temporally sequenced in the order that they occur, which may be different than the order in which they are told. 
Most notably, the events and facts in the fabula might not all exist in the final telling of the narrative; some events and facts might need to be inferred from what is actually told.
It is not required that the syuzhet to be told in chronological order, allowing for achronological tellings such as flash forward, flashback, ellipses (gaps in time), etc.

% Some narratologists provide a third layer of analysis, the text, which is outside the scope of this paper. 
They key is that readers interact more closely with syuzhet and must infer the fabula through the text of the syuzhet.
Because a fabula inferred, it may be occuring in one of many possible worlds in a modal logic sense~\cite{ryan1991possible}. 

\begin{definition}
A {\bf story world} is a set of possible worlds that are consistent with the facts and events presented to the reader in the syuzhet.
\end{definition}

\noindent
As events and facts are presented throughout the narrative, the probability cloud over story worlds collapses and a reader's beliefs become more certain.
% Having defined a notion of the relevance of Question $A$ to Question $B$, our next step is connecting to existing narratological analysis. Consider Barthes' notion of kernels and satellites. \cite{barthes1975introduction}

Events in the fabula and story world have different degrees of importance:

\begin{definition}
    A \textbf{kernel} is a narrative event such that after its completion, the beliefs a reader holds as they pertain to the story have drastically changed.
\end{definition}

\begin{definition}
    A \textbf{satellite} is a narrative event that supports a kernel. They are the minor plot points that lead up to major plot points. They do not result in massive shift in beliefs.
\end{definition}

\noindent
Satellites imply the existence of kernels, e.g. small plot points will explain and lead up to a large plot point, but kernels do not imply the existence of satellites---kernels do not require satellites to exist. 
%One can think of this as when satellites exist kernels must always exist on their boundary whether they are referred to in the text or not.
%
A set of satellites, $s = \{s_1, \hdots, s_n\}$, is said to be relevant to a kernel $k$ if, after the kernel's competition, the reader believes that the set of questions posed by $k$ are relevant to their understanding of the story world given prior $s$.

An implication of kernels and satellites is that one can track a reader's understanding of a story over time by asking the reader questions relevant to the story before and after each major plot point.
As kernels change the reader's beliefs about the story world and the fabula, then their answers to questions change as well.

%%%%%%%%%%%%%%%%%%%%%%%%%%%%%%%%%%%%%%%%%%%%%

\section{Fabula Entropy Indexing}

%\Mark{This section should give the definitions of ETC and EWC, making reference to preliminaries. It can't reference line numbers or when questions are asked. Because those haven't been introduced yet. Most importantly, this section needs to give the human participant methodology---questions get generated, questions get asked, entropy is computed.}

Fabula Entropy Indexing (FEI) measures story coherence based on human question-answering.
Humans build cognitive structures to make sense of a story, which in turn can be used to answer simple true/false questions about the story.
A coherent narrative results in readers having well-formed cognitive models of the fabula and story world\cite{graesser2003readers,trabasso1982causal}.
Because the cognitive models formed during reading are predictable across readers
one can infer that coherent stories result in readers being more likely to answer questions about a story similarly~\cite{graesser91}.
{\em Incoherent} stories thus result in readers making random guesses about the answers to questions.
FEI looks at the entropy of the answers---how much readers disagree with each other---as a signal of coherence of the story.

We decompose FEI into two separate metrics. 
{\em Entropy of Transitional Coherence} (ETC) 
measures the necessity of transitional ordering: in time $t$, event or fact $x$ is necessary to maintain a story's coherence.
In other words, was this fact probable before $t$? 
This establishes whether a reader could reasonably anticipate the occurring between two events. 
{\em Entropy of World Coherence} (EWC)
on the other hand is not time dependent. 
EWC measures the probability of an event or fact $y$ occurring {\em at any time} in a story world.

%\MarkRight{Move later, entropy not introduced yet}
%Since relevance is defined as the difference of entropy between two queries, we can measure disagreement between human participants as the average relevance of a query when compared to some notion of ground truth, conditioned on the kernel. 

%\Mark{Need transition to set up entropy}
The core idea of Fabula Entropy Indexing is that readers can be asked true/false questions and that the agreement in readers' answers indicates coherence.
However, questions must take the form of implications $q:A\implies B$ (read ``if $A$ then $B$'') and the two propositions $A$ and $B$ must have {\em relevance} to each other.

%Before delving into the technical details of our metrics, we must first understand what it means for the answer of question to be relevant to the answer of another. 

\begin{definition}
For a question about a story, $q$, of the form ``if A then B'' with possible values for $A=\{T, F\}$ and possible values for $B=\{T,F\}$.
 Identifying $A$ with the set of possible answers to it, we say that the {\bf relevance} of $B$ to $A$ given some prior $\gamma$ is 
    \begin{equation}
    H(A = a_i | \gamma) -  H(B = b_j | A = a_i, \gamma) 
    \end{equation}
    where $a_i$ and $b_j$ are the true answers to $A$ and $B$ and $H$ refers to binary entropy. \cite{castricato2021towards}.
\end{definition}

\noindent
Note that the relevance of $B$ to $A$ \textit{depends on the ground truth}.
%This is perhaps surprising, but after some consideration it should be clear that this has to be true. After all, the causal relationship between $A$ and $B$ could depend on the true answers! 
Consider the case where $A$ is ``is Harry Potter the prophesied Heir of Slytherin?'' and $B$ is ``can Harry Potter speak Parseltongue because he is a descendent of Slytherin?''. 
If Harry is a blood descendant of Slytherin and that is why he can speak Parseltongue, then $B$ is highly relevant to $A$. 
However, the actual truth of the matter is that Harry's abilities are completely independent of his heritage. 
%and arose due to a childhood experience. 
Therefore $B$ does not have relevance to $A$ even though \textit{it could have had relevance} to $A$ had the ground truth been different.

\subsection{Entropy of Transitional Coherence}

Certain facts or events in stories have temporal dependencies. 
For example, a protagonist may hammer a nail into the wall. If subsequent events reveal the fact that the protagonist never held a hammer this causes temporal or transitional incoherence.

If we force our question to be an implication, namely of the form ``Given that $A$ occurs within the story, then $B$", 
we are attempting to determine the relevance of a query $B$ to a query $A=true$, specifically: $$H(A = true | \gamma) -  H(B = b_j | A = true, \gamma).$$
If $A$ is given within the reader's inferred fabula, then $A$ is always true and we simply want to query about $B$. However if $A$ is undetermined within the reader's inferred fabula then we are as a whole querying about ``If $A$ then $B$,'' and forcing the reader to reconcile both $A$ and $B$ without any belief about $A$. 

Entropy of Transitional Coherence therefore asks questions of readers in which $A$ is a belief from before a kernel and $B$ is a belief from after a kernel.
Let question $q$ be of the form 
``Given that $A$ occurs within the story, then $B$.''
That is $q := A \implies B$.
Let $P(q)$ refer to the proportion of story worlds where $q$ is true.
The stronger the reader's belief, the more possible worlds in which $q$ is true, and the higher the probability.
Across all readers answering the question:
\begin{equation}
\begin{aligned}
H&(P(q))  = H(q | \gamma) \\
& = H(A = T | \gamma) - H(B = b_j | A = T, \gamma)
\end{aligned}
\end{equation}

% If $A$ is a belief before a kernel and $B$ is a belief after a kernel, then we are determining the agreement of a reader's internalized transition model of the story world. This now brings us to ETC. Let $Q$ refer to a set of statements of the form ``Given that $A$ occurs within the story, then $B$." For $q \in Q$, let $P(q)$ refer to the proportion of story worlds, defined by $\gamma$, where $q$ is True. It can be shown that if $q := A \implies B$,
% $$H(q | \gamma) = H(A = T | \gamma) - H(B = b_j | A = T, \gamma)$$ 

% \noindent
% Therefore, since by definition $P(q)$ is conditioned on $\gamma$, we can average the entropy given above over all such $q \in Q$ and arrive at

By averaging across all questions $Q$ that span kernels, we arrive at the definition of ETC:

\begin{equation}
\label{eqn:entropy}
E(Q) = \frac{1}{|Q|}\sum_{q \in Q} H\big(P(q)\big)
\end{equation}

\noindent
In the context of Entropy of Transitional Coherence, $ETC(Q) = E(Q)$.

Consider the following example for discussing the importance of ETC. A person needed a bath, so they went for a run. A possible query here would be ``Given a person needed a bath, does this contradict that they went for a run?" In this particular example, we can assume going for a run is a kernel and as such this query measures if needing a bath is a plausible precondition to desiring to go on a run. Equivalently, does the reader believe ``If the person needs a bath, then they go for a run.'' If the story makes less sense to the reader, the reader attempts to reconcile these two clauses and as such would be more likely to guess.\cite{trabasso1982causal,mandler1977remembrance}

\subsection{Entropy of World Coherence}

Whereas Entropy of Transitional Coherence measures coherence as events cause the story world to change, Entropy of World Coherence (EWC) measures the coherence of static fact about the story world. 
For example if a story contains a protagonist that is described as being short but is also described as hitting their head on the top of a doorframe, we might find readers have more varied responses to a question about the protagonist's height.

Entropy of World Coherence also uses Equation~\ref{eqn:entropy} (that is, $EWC(Q)=E(Q)$) but does not require that the questions reference before and after kernels.
There need not be any temporal requirement to questions.
Instead EWC relies on questions about descriptive elements in a story, as signified by adjective and adverbs.
However, these descriptions of characters, objects, or places must be integral to at least one event in the narrative.

\subsection{Measuring Coherence with Human Participant Studies}
\label{sec:measure}

Having mathematically defined our two coherence metrics, ETC and EWC, as a function of readers responding to a set of questions about temporal or non-temporal aspects of a story,
we now describe how we use ETC and EWC to measure coherence of stories, particularly those from by automated story generation systems.
%
%The main purpose of Fabula Entropy Indexing is to evaluate the coherence of automated story generation systems.
There are three key steps to Fabula Entropy Indexing as a methodology.

The first step is to use an automated story generation system to generate a number of stories that are representative of its capabilities. 
Typically this would be done by randomly seeding the generator.

The second step is to produce a number of questions. 
To produce questions for ETC, one identifies the kernels---the major plot points---and constructs questions such as:

\begin{itemize}
    \item Does Entity A's sentiment/emotion change between line N-1 and N?
    \item Does Object A change possession in Line N+1?
\end{itemize}

\noindent
To produce questions for EWC, one identifies adjectives and adverbs that could be changed, such as: 

\begin{itemize}
    \item Does [Adverb/Adjective] contradict an assertion on Line N?
    \item Could [Adverb/Adjective] be removed and the story world would remain unchanged?
\end{itemize}

\noindent
One would want to produce as many questions as possible. Note that while the questions above do not read as implications immediately, they can be expressed as the required implications after a bit of work and thus still satisfy our constraint.

It doesn't matter what the questions are or what the answers are---we do not require a ground truth---as long as the questions reference aspects of the story that can impact readers' cognitive model formation.
ETC and EWC guide us toward kernels and attributes, respectively.
Fabula Entropy Indexing measures coherence by observing the agreement between human participants when answering these questions.

The third step is to recruit human study participants to read a story and then answer the associated questions.
There is no ground-truth ``correct'' answers---we are not testing participants ability to answer in a certain way.
Instead, we use Equation~\ref{eqn:entropy} to measure agreement between responses, under the assumption that more coherent stories prompt readers to construct more consistent mental models of the fabula and story world. 

ETC and EWC can be compared between representative sets of stories between different automated story generation systems.
Lower entropy values implies greater coherence.

% To construct our measure, we first source short stories from a public repository. Short stories are desirable as human participants will need to answer questions about the interdependence between all plot points and simplifies the task to make it more suitable for crowdsourcing. There are three separate human participant tasks, each completed by a new set of human participants. The three tasks are 1) alter a story in such a way that it introduces improbable events or facts, 2) generate questions about a story from a set of templates \Spencer{Need justification for constructing the templates ourselves?} and 3) answer the questions generated by participants in the second task. Separate participants view a mix of ground truth and corrupted stories but never both the corrupted and uncorrupted version of a story. For ETC, human participants do not corrupt these stories - the process of rearranging plot points is random and automated. By answering these questions, it is possible to compare readers' belief in the possibility of a corrupted event versus the events in the unchanged story. 

%%%%%%%%%%%%%%%%%%%%%%%%%%%%%%%%%%%%%%%%%%%%%%%%%%%%%%%%%%%%%%%%%%%%%%%%%%%%%%%%%%%

\section{Experiments}

To validate Fabula Entropy Indexing in general, and ETC and EWC in particular, we need to verify that the methodology in Section~\ref{sec:measure} produces low entropy values for coherent stories and high entropy values for incoherent stories.
Because automated story generation is still an open research question, we validate ETC and EWC on human-written stories that are known to be coherent.
We assume that human-written stories are coherent.
To compare entropy indices against incoherent stories, we devise a technique for corrupting human written stories in particular ways that are likely to result in incoherent stories. Exemplar corruptions include negating adjectives, swapping events from different stories or randomly changing key descriptors of characters.

% \subsection{Methods}

% Due to the nature of ETC and EWC, we test them in very controlled settings. We first approached the task of creating artificial data as a process of taking known coherent stories and corrupting them in various ways which we will outline below.

\subsection{Entropy of World Coherence Stories} 

For EWC, we source a number of short stories by authors such as Rumi, Tolstoy and Gibran. Specifically, this is a subset available in a public repository\footnote{https://github.com/pelagia/short-stories} unaffiliated with the authors of this paper. For each story we subdivide them into 10-line segments if the story was longer than 10 lines. 
We selected 9 stories for the experiment.\footnote{In both the ETC and EWC cases we had intended to evaluate over 10 stories but one story was rejected due to one of the stories inadvertently having a controversial interpretation when corrupted and which was only pointed out to us by one of the question-answering participants.}

To create a corrupted story baseline in which story coherence is less assured, we copied the 9 stories and made changes to them.
We recruited 4 participants who are unaffiliated with the research team and asked them to independently select a subset of the adjectives and adverbs from a story and swap them for their antonyms. 
This produced stories that are, at a story world level, less coherent since due to the highly descriptive nature of the stories one swap was more likely to lead to a contradiction later on in the story. Participants were required to create the inconsistency and not to fix their incoherency with more swaps.
Participants were compensated \$20/hr to complete this task.

\subsection{Entropy of Transitional Coherence Stories} 

For Transitional Coherence we require a direct correspondence between events and sentences. 
{\em Plotto}~\cite{cook2011plotto} is a compilation of plot points with annotations about which plot points can be followed by others.
%It can be used to generate plot outlines. 
Plotto can thus be used to generate plot outlines assembled from human-written segments. 
The Plotto plot points contain few adjectives and plot outlines generated from the Plotto technique are unambiguous with respect to transitions in the story world. Since plotto consists of plot points, every vertex, and in our case line number, using the Plotto technique is a kernel. Within every kernel are a number of sentences, typically 2-3, that denote the satellites.

Since Plotto directly states plot points rather than having the reader infer them, this allows us to controllable corrupt the order of plot points by swapping lines- something that is rarely possible with human written short stories.

To construct stories for measuring ETC, we use the Plotto technique to generate 5-6 sentence short stories. 
%We follow the Plotto technique to create non-corrupted stories.
For the experiment we generated 9 stories in this way.

To construct corrupted stories, we copied the 9 stories above and then swap the order of plot points, which results in incoherence (e.g. a burglar getting away with a crime before they're even born).
We generate Plotto stories with 5 vertices, and randomly choose a span of 3 vertices. Within that span, we shuffle their order.

% and ask Plotto to expand the narrative graph.

% Secondly, for ETC, we took extract plot graphs to generate 5-6 sentence short stories from Plotto \cite{cook2011plotto}, which are known to have few adjectives and be unambiguous with respect to transitions in the story world. Plotto first uses algebraic rules to construct a narrative graph and then within each vertex (sentence or plot point) expands the story further. Because of this, we can swap the order of vertices in our narrative graph to introduce temporal incoherence within a story (e.g.a burglar getting away with a crime before they're even born.)

% We generate Plotto stories with 5 vertices, and randomly choose a span of 3 vertices. Within that span, we shuffle their order and ask Plotto to expand the narrative graph.

\subsection{Question Generation}

To measure ETC and EWC we require a set of true/false questions for each story.
To ensure that we do not introduce experimental bias in questions for each story, we recruited 4 people to write questions for each story.
Question writers were compensated \$20/hr and produced 10-15 questions per story.

% \Mark{Depending on what section \ref{sec:measure} says, we may need to talk about question templates here. Are templates part of the methodology for measuring ETC and EWC or are does the measure assume that questions exist and that they are experimental methodology details particular to the stories (and thus to be described here)?} \Spencer{I made mention of the templates in measure but agree there needs to be a discussion about the protocol used to construct the templates or at least better justification for their use (since we constructed the templates)}

%\Spencer{Assume we will include all templates and questions in the appendix? And the "Task Packet" content/instructions?}\Mark{Maybe just the full set of templates are necessary.}
%\Mark{Why does this only talk about corrupted EWC stories?}
For the corrupted sets of both Plotto and non-Plotto stories, we task a human participant to write questions guided by a set of templates which provide the best coverage over the more likely reader possible worlds. That is to say, if there were N reasonable interpretations of the story, we aimed to have our human subjects construct questions that could differentiate between N interpretations. Said another way, all templates probe the probability or plausibility of one plot point occurring or impacting the reader's comprehension of other plot points, in some way. 

Participants were provided a packet which includes a description of the research, instructions for the task and a list of templates to follow when generating questions. Templates were also used to standardize the format of questions human participants in the subsequent experiment would receive. Question writing participants could freely choose the entities, properties and line numbers represented in each question.

% ETC does not require human input as aforementioned, the process of temporal corruption can be automated and random.

% ETC questions were generated in the same guided manner by tasking a human participant to write a series of template-driven questions.
% We only constructed these questions for the corrupted set since ... (it was possible to swap the corruptions and reuse, for now) \Louis{I forgot why.}

A partial list of corruption prompts and a full list of question templates with some exemplar completions are provided in the Appendix.

\subsection{Methodology}

For each task, we recruit 180 participants on the Prolific platform, split evenly between ETC and EWC tasks. Demographic screening excluded any non-US individuals, individuals for whom English is not their first language, as well as those with linguistic impediments on the basis of the tasks' relative comprehension complexity. Each worker was either given corrupted stories or uncorrupted stories, but never both. This was done to prevent a worker from seeing both the uncorrupted and corrupted version of a story and as such biasing the results. Every worker received a randomized set of 3 stories. For each story, 10-15 yes or no questions were asked about interdependencies between sentences of the same story. Workers were compensated \$20/hr for their time and given a screening question that was a handmade EWC and ETC example respectively. These examples were not used in computing the final result.

\subsection{Results}

\begin{figure}[t]
\centering
\includegraphics[scale=0.3]{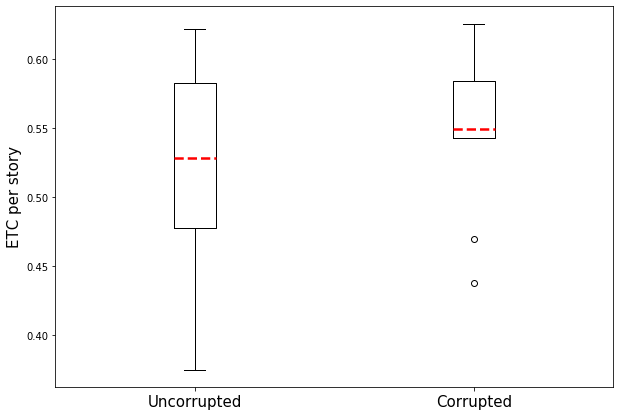}
\caption{Entropic indices of transitional coherence derived from human participant evaluation of Plotto stories. Lower is better.
% evaluation shows that \sysname{} scores 0.444 on the entropy index, compared to Backwards BART's 0.520. Human written stories, ROC, scores a 0.26, over 50\% better than Backwards BART and over 41\% better than EDGAR. 
% Notice mninimal overlap between the interquartile ranges.
}
\label{fig:etc}
\end{figure}

\begin{figure}[t]
\centering
\includegraphics[scale=0.3]{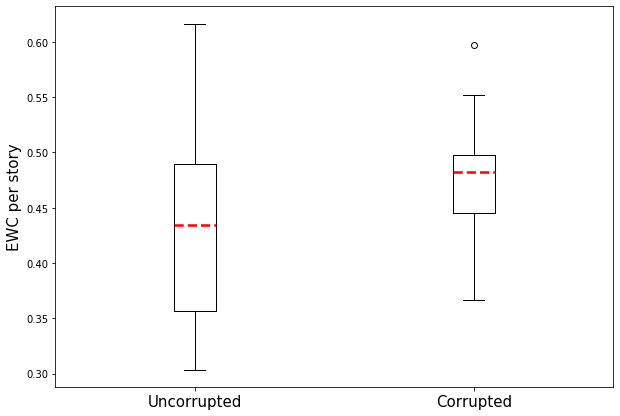}
\caption{Entropic indicies of world coherence derived from human participant evaluation of the non-Plotto story dataset. Lower is better.
% evaluation shows that \sysname{} scores 0.444 on the entropy index, compared to Backwards BART's 0.520. Human written stories, ROC, scores a 0.26, over 50\% better than Backwards BART and over 41\% better than EDGAR. 
% Notice mninimal overlap between the interquartile ranges.
}
\label{fig:ewc}
\end{figure}

The results are summarized in Figure~\ref{fig:etc} for Entropy of Transitional Coherence and Figure~\ref{fig:ewc} for Entropy of World Coherence. The bars on the left are the results for uncorrupted, original stories and the bars on the right are for the stories modified to corrupt coherence.
The red line indicates the mean of each distribution. Median is not reported.
The results suggest that original stories have lower entropy and are thus more coherent.
This validates fabula entropy indexing because the corruptions we applied to the same set of stories are designed to interfere with readers' abilities to form a well-formed model of the fabula and story world. 
% When answering questions about plausibility and possibility for temporal and non-temporal plot interdependencies, participants indicated the corrupted stories were less coherent significantly more frequently.

We do not report statistical significance because statistical significance tests are undefined on entropy distributions, which are not probability distributions.

%%%%%%%%%%%%%%%%%%%%%%%%%%%%%%%%%%%%%%%%%%%%%%%%%%%

\section{Discussion}

From the results, we can make some observations.
The first is that the corrupted stories are not a traditional experimental baseline.
The corruptions were designed to show that intentionally introduced incoherencies do in fact result in an increase in entropy.
Second, the corruptions are designed to introduce the smallest possible amount of incoherence to stories as possible.
Therefore, we would not expect a large increase in entropy due to a single corruption per story.
The fact that entropy increases with the introduction of minimalist corruptions indicates that Fabula Entropy Indexing is sensitive to such small changes.
We would anticipate an automated story generator that routinely makes transitional or world coherence errors to result in much more significant differences in entropy values.

The entropies for corrupted stories have more dense distributions.
Not only was there more disagreement about the answers to questions, but the disagreement was consistent across all stories. 
This is to be expected because the corruptions are synthetically designed to damage story coherence.
The entropy distributions for real stories was spread over a wider range of entropy values per story.

ETC might not be as strong a metric as EWC.
The average ETC of uncorrupted stories is higher than the EWC of uncorrupted stories. 
This may be due to (a)~human tolerance for event ordering variations; (b)~the Plotto technique may have produced plots in which plot points are only loosely connected; (c)~our swap-based corruptions may not always produce incoherent stories.

The quality of the entropy indices are highly dependent on the extent to which the true/false questions target points in the story where potential incoherence can arise.
It may theoretically be possible for some automated story generators to automatically generate good sets of questions, however this is currently an open research problem. 
The authors of this paper could have generated a better set of true/false questions targeting ETC and EWC than those unaffiliated with the research.
However, doing so introduces the possibility of experimenter bias, which needs to be avoided by those who use this evaluation technique.

FEI has a couple of limitations.
First, to measure ETC one must be able to identify kernels and make questions about elements before and after the kernels.
Second, to measure EWC, the stories must be highly descriptive in nature and that there are plot points that are dependent on adjectives; many story generators do not produce descriptive texts.

FEI was validated on short stories, of 10 sentences or less.
While there is no theoretical reason it will not work on longer stories, it will require substantially more questions to be produced and answered by human participant studies.

We have used the Fabula Entropy Indexing method described in this paper to evaluate an automated story generation system in (under review, 2021).
The REDACTED system was designed explicitly to increase coherence of automatically generated stories over a large pretrained transformer language model baseline. 
The combined ETC and EWC for the experimental system were lower than the language model baseline. 
Moreover, we also compared the entropy indices of human-written baseline stories, showing that human stories result in lower entropy values than AI generated stories, which is to be expected at this time.
This constitutes the first successful use of FEI for its intended purpose of evaluating automated story generation systems.

As part of the above real-world test case of FEI, we also performed a {\em subjective} human-participant study, showing that the entropy indices are low when humans report perceived coherence.
We did not perform a subjective human participant study for this paper since we were working on stories that came from sources with reliable coherence.

%%%%%%%%%%%%%%%%%%%%%%%%%%%%%%%%%%%%%%%%%%%%%%%%%

\section{Conclusions}

Automated Story Generation research requires strong, reliable evaluation metrics, which have largely been absent, hampering research progress.
We present the Fabula Entropy Indexing technique for objectively evaluating the coherence of stories. We demonstrate the effectiveness of this technique by showing how two FEI metrics, entropy world coherence and entropy transitional coherence, can be used to clearly discriminate between stories with and without coherence corruption. 
In contrast to subjective human participant studies, where it is challenging to get participants to answer questions about coherence, FEI provides a numerical rating of the coherence of stories that is grounded in theory.

%%%%%%%%%%%%%%%%%%%%%%%%%%%%%%%%%%%%%%%%%%%%%%%%

\typeout{}
\bibliographystyle{acl_natbib}
\bibliography{naacl_applied}
\appendix

\section{Appendices}

\subsection{Alteration Templates\footnote{Additional clarifying examples were given to participants when they requested them during task completion.}}

The [Adjective1] Object/Entity/Event -> The [Adjective2] Object/Entity/Event\\
\\
The [Adjective1] Object/Entity/Event -> The not [Adjective1] Object/Entity/Event\\
\\
Object/Entity/Event is [Adverb1] [Adjective1] -> Object/Entity/Event is [Adverb1] [Adjective2]\\
\\
Object/Entity/Event is [Adverb1] [Adjective1] -> Object/Entity/Event is [Adverb2] [Adjective1]\\
\\
Object/Entity/Event [Adverb1][Verb] -> Object/Entity/Event [Adverb2][Verb]\\

These are just a small sample of templates given the complex nature of certain sentences. You can make alterations beyond this but adhere to the rules above.\\

\subsection{Question Templates: EWC}

In the context of this narrative setting, is [Adverb/Adjective] plausible? (e.g. an ``otherworldly'' dog showing up in a short story about World War 2 where you might otherwise describe a ``stray'' dog. Note: This may not be a constraint for all readers - those answering questions will only assess based on {\em their} belief about the world.)\\
\\
Prior to this line did you imagine [Adverb/Adjective] was a possible descriptor for Object/Entity/Event?\\
\\
After this line containing [Adverb/Adjective] do you hold the belief this is a possible descriptor or do you reject it?\\
\\
Because of [Adverb/Adjective] does Line N contradict information in another line?\\
\\
Because of [Adverb/Adjective] does this indicate emotional valence (extreme sentiment) toward an Object/Entity/Event?\\
\\
In the line with [Adverb/Adjective] does this alter Author or Entity sentiment toward Object/Event?\\
Because of [Adverb/Adjective] does this change your sentiment toward some Entity/Object/Event?\\
\\
Does [Adverb/Adjective] contradict an assertion on Line N?\\
\\
Could [Adverb/Adjective] be removed and the story world would remain unchanged?\\
\\
Without [Adverb/Adjective] on Line N, Line N+1 would not have happened.\\

\subsection{Question Templates: ETC}

Does Entity A's perception of Entity B change?\\
\\
Do all Entities in Line N observe or gain awareness of Events in Line N+1?\\
\\
Do the Events in Line N+1 contradict Events in Line N?\\
\\
Does Entity A's sentiment/emotion change between line N-1 and N?\\
\\
Does Object A still retain State S?\\
\\
Does Object A change possession in Line N+1?\\
\\
Is Object A in Line N+1 necessary for Events in line N to occur?\\
\\
Is there a change in context or location between these lines?\\
\\
Is knowledge of Object A necessary for understanding the following line?\\
\\
Does Line N have causal dependencies established in Line N-1?\\
\\
Could Line N-1 occur before Line N?\\

\subsection{Selected Questions}

Does "awful" contradict an assertion on line 1?\\
\\
Could "shaped" in line 4 be removed and the story world would remain unchanged?\\
\\
Because of "tall" does line 9 contradict information in another line?\\
\\
Could line 1 and 5 both be removed and have no effect on the story?\\
\\
Is there a change in context or location between line 2 and 5?\\
\\
Do the events in line 3 contradict events in line 2?\\
\\
\end{document}